\documentclass[letterpaper]{article} % DO NOT CHANGE THIS
\usepackage{aaai2026}  % DO NOT CHANGE THIS
\usepackage{times}  % DO NOT CHANGE THIS
\usepackage{helvet}  % DO NOT CHANGE THIS
\usepackage{courier}  % DO NOT CHANGE THIS
\usepackage[hyphens]{url}  % DO NOT CHANGE THIS
\usepackage{graphicx} % DO NOT CHANGE THIS
\usepackage{natbib}  % DO NOT CHANGE THIS AND DO NOT ADD ANY OPTIONS TO IT
\usepackage{caption} % DO NOT CHANGE THIS AND DO NOT ADD ANY OPTIONS TO IT
\usepackage{algorithm}
\usepackage{amsmath}
\usepackage{amssymb}
\usepackage{amsfonts}
\usepackage{multirow}
\usepackage{booktabs}
\usepackage{subcaption}
\usepackage{algpseudocode}

\usepackage{newfloat}
\usepackage{listings}
\DeclareCaptionStyle{ruled}{labelfont=normalfont,labelsep=colon,strut=off} % DO NOT CHANGE THIS
\floatstyle{ruled}
\newfloat{listing}{tb}{lst}{}
\floatname{listing}{Listing}
\title{HQ-OV3D: A High Box Quality Open-World 3D Detection Framework based on Diffision Model}

\author {
    Qi Liu\textsuperscript{\rm 1,\rm 2,\rm 3}\thanks{Equal contributions.}
    Yabei Li\textsuperscript{\rm 4}\footnotemark[1]
    Hongsong Wang\textsuperscript{\rm 3}
    Lei He\textsuperscript{\rm 1,\rm 2}\thanks{Corresponding author: helei2023@tsinghua.edu.cn}
}

\affiliations{
    \textsuperscript{\rm 1}School of Vehicle and Mobility, Tsinghua University, Beijing 100084, China.\\
    \textsuperscript{\rm 2}State Key Laboratory of Intelligent Green Vehicle and Mobility, Tsinghua University, Beijing 100084, China.\\
    \textsuperscript{\rm 3}School of Computer Science and Engineering, Southeast University, Nanjing 210096, China\\
    \textsuperscript{\rm 4}Meituan Inc.\\
}
\begin{document}
\makeatletter
\maketitle
\makeatother
\begin{abstract}
%Open-vocabulary 3D object detection for autonomous driving has emerged as a prominent research topic in the field of open-world perception.
Traditional closed-set 3D detection frameworks fail to meet the demands of open-world applications like autonomous driving.
Existing open-vocabulary 3D detection methods typically adopt a two-stage pipeline consisting of pseudo-label generation followed by semantic alignment. While vision-language models (VLMs) recently have dramatically improved the semantic accuracy of pseudo-labels, their geometric quality, particularly bounding box precision, remains commonly neglected.
%The pseudo-label generation process in these approaches heavily relies on hand-crafted heuristic-based constraints. As a result, the generated pseudo-labels often suffer from limited accuracy and high noise, severely hindering downstream modeling capabilities and detection performance.
To address this issue, we propose a High Box Quality Open-Vocabulary 3D Detection (\textbf{HQ-OV3D}) framework, dedicated to generate and refine high-quality pseudo-labels for open-vocabulary classes. The framework comprises of two key components: a \textbf{Intra-Modality Cross-Validated} (IMCV) Proposal Generator that utilizes cross modality geometric consistency to generate high quality initial 3D proposals, and a \textbf{Annotated-Class Assisted} (ACA) Denoiser that progressively refine 3D proposals by leveraging geometric priors from annotated categories through a DDIM-based denoising mechanism. Compared to the state-of-the-art method, training with pseudo-labels generated by our approach achieves a \textbf{7.37\%} improvement in mAP on novel classes, demonstrating the superior quality of the pseudo-labels produced by our framework. 
%HQ-OV3D can serve as a strong upstream data generation module, offering more reliable pseudo-label support for subsequent semantic alignment and open-vocabulary tasks.
HQ-OV3D can serve not only a strong standalone open-vocabulary 3d detector but also a plug-in high quality pseudo-label generator for exiting open-vocabulary detection or annotation pipelines.
\end{abstract}


\begin{figure}[htbp]
    \centering
    \begin{subfigure}[b]{0.45\linewidth}
        \includegraphics[width=\linewidth]{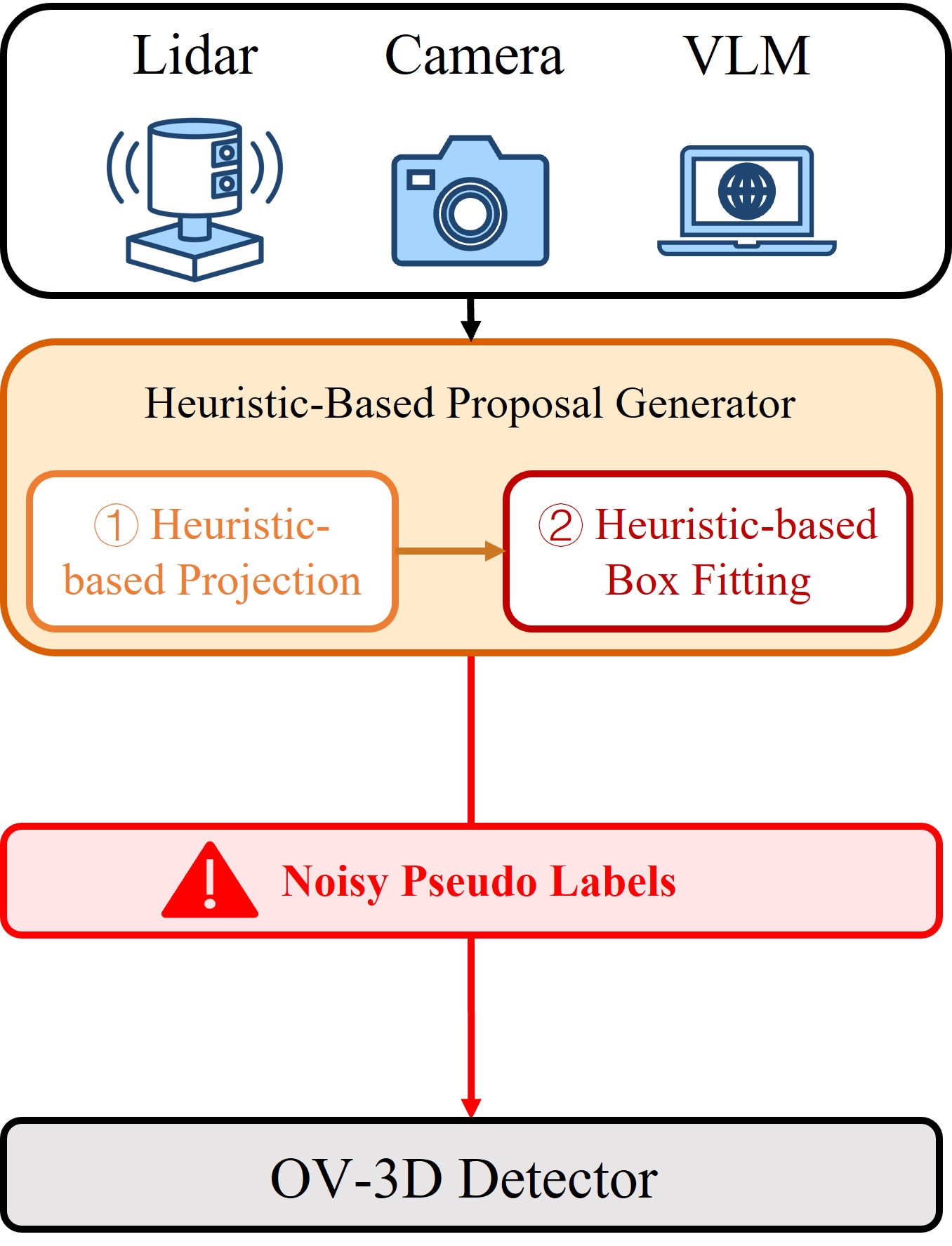}
        \caption{Existing OV-3D appoach}
        \label{fig:Introduction1a}
    \end{subfigure}
    \hspace{0.05\linewidth}
    \begin{subfigure}[b]{0.45\linewidth}
        \includegraphics[width=\linewidth]{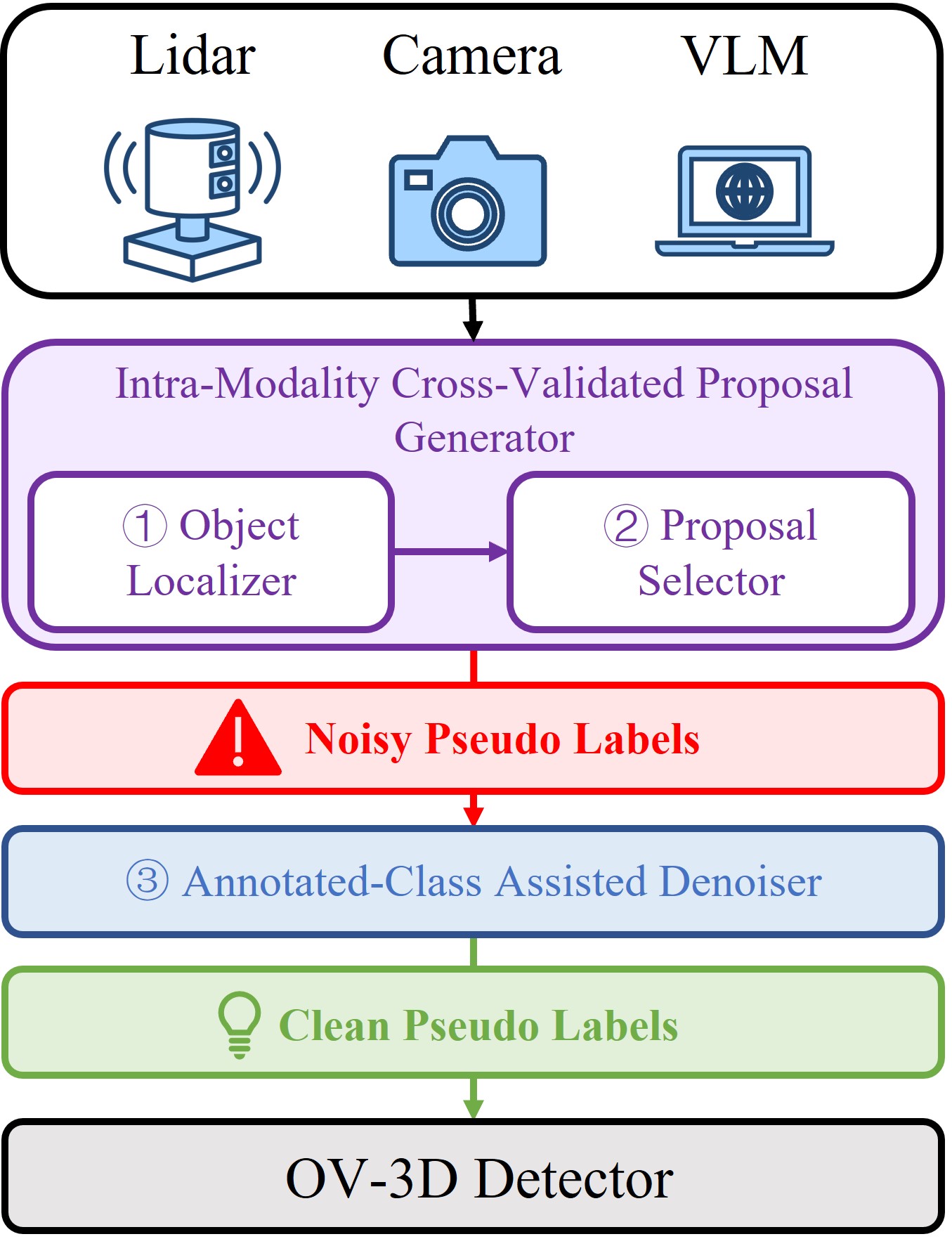}
        \caption{Our proposed HQ-OV3D approach}
        \label{fig:Introduction1b}
    \end{subfigure}
    \caption{Comparison between existing OV-3D pipeline and our proposed HQ-OV3D pipeline. (a) Existing methods use heuristic-based projecting and box fitting, which generates low-quality pseudo-labels with noise. (b) Our method proposes the IMCV Proposal Generator and ACA Denoiser to generate high quality box for OV-3D.}
    \label{fig:Introduction1}
\end{figure}

\section{Introduction}
Recent years have witnessed significant advancements in technologies like autonomous driving, embodied intelligence, etc. However, as these systems are increasingly deployed in open-world scenarios, they frequently encounter diverse corner cases. Long-tail open-set 3D object perception has emerged as a fundamental bottleneck, restricting their wider applications.
Traditional closed-set 3D object detection pipelines require retraining from scratch whenever novel categories are introduced, a process that demands extensive collection of new 3D bounding box annotations and substantial computational resources, making it both time-consuming and labor-intensive. Recently, the Open-Vocabulary 3D Object Detection (OV-3D) task has been introduced, aiming to empower models with the ability to recognize and localize arbitrary novel semantic categories.

\begin{figure}[t]
\centering
\includegraphics[width=0.45\textwidth]{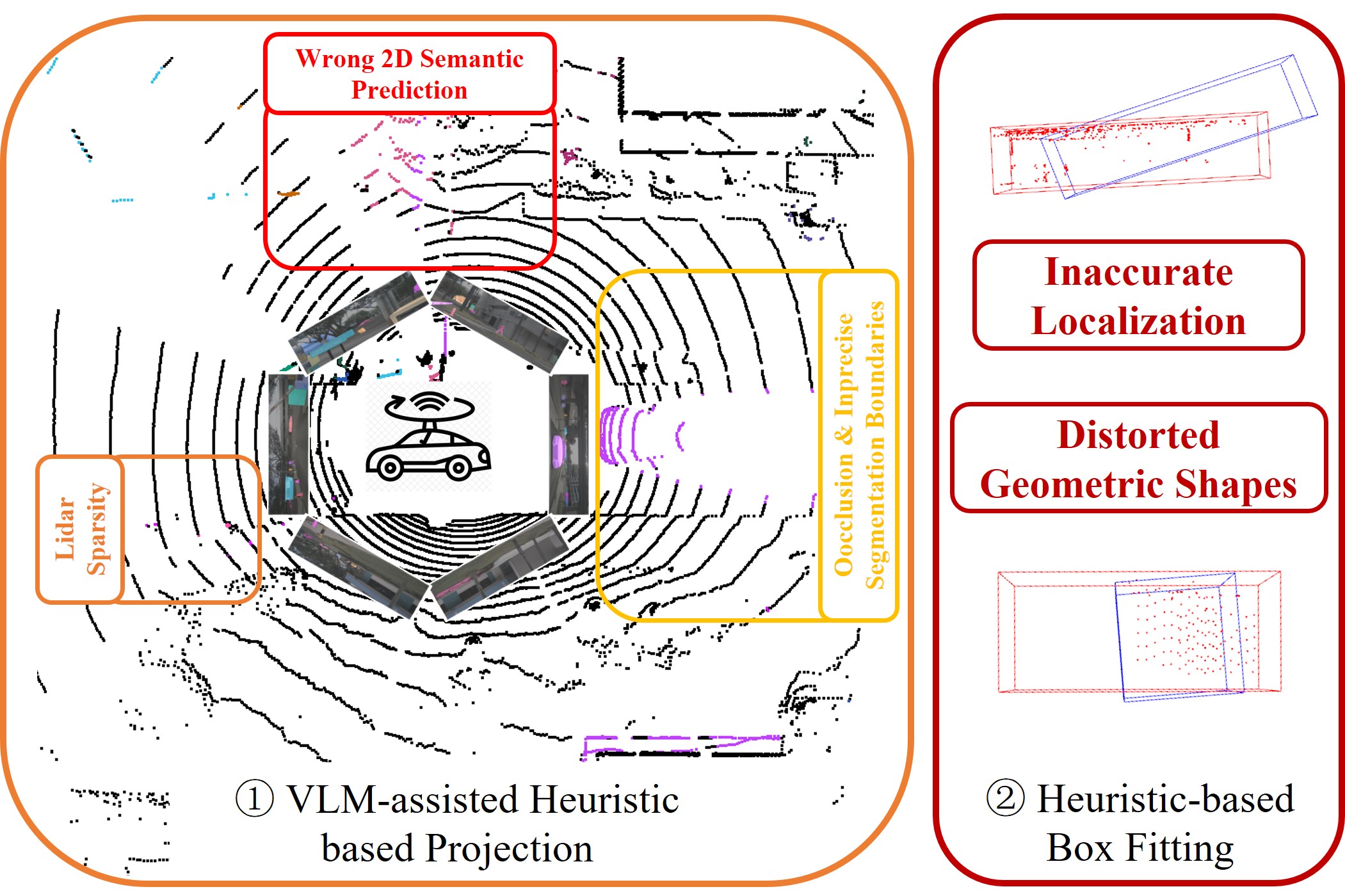}
\caption{Illustration of the limitations of heuristic-based pseudo-label generation in OV-3D. These limitations motivate the need for a more robust pseudo-label generation and refinement framework.}
\label{fig:Introduction2}
\end{figure}

Inspired by the remarkable open-set semantics capability of vision-language models (VLMs) in image understanding, most current OV-3D approaches~\cite{zhang2024opensight,chow2025ov,cao2023coda}adopt a two-stage paradigm: \textit{pseudo-labels generation} followed by \textit{semantic alignment}. In the first stage, 3D pseudo-labels are generated by heuristic projecting VLM-derived 2D perception results to 3D space through geometric priors and heuristic box fitting. In the second stage, VLM-derived semantic embeddings or textual prompts are used to align point cloud features of 3D pseudo-labels, improving the model's capacity to recognize novel classes. Despite the effectiveness, the framework suffers from two key limitations. Firstly, the box quality of generated pseudo-labels is overlooked. Projected from 2D VLM prediction, 3d pseudo-labels suffer from geometrical misalignment (such as depth ambiguous, occlusion), noisy perception inherited from 2D prediction (such as imprecise segmentation boundaries and semantics), lidar sparsity, as illustrated in Figure~\ref{fig:Introduction2}. The subsequently box fitting pipeline could only rely on limited geometrical info in sparse, possibly incomplete  point cloud. Consequently, the generated pseudo boxes exhibit two critical imperfections: (1) inaccurate localization (2) distorted geometric shapes, as shown in Figure~\ref{fig:Introduction2}. These defects severely degrade downstream cross-modal feature alignment and systematically propagate errors to the final open-vocabulary 3D detection results. Secondly, known categories annotated boxes are underutilized. In practice, novel categories usually emerge incrementally without any annotations while known categories such as vehicles and pedestrians usually have already maintain rich annotations. Annotated boxes could offer not only semantic labels but also rich structural and geometric priors, which can be transferred to improve novel categories' detection box precision.

To address the above limitations, we present \textbf{HQ-OV3D} (High-Quality Open-Vocabulary 3D Detection), a novel framework designed to improve bounding box quality through two key components: the \textit{Intra-Modality Cross-Validated Proposal Generator} (IMCV Proposal Generator) and the \textit{Annotated-Class Assisted Denoiser} (ACA Denoiser). Figure~\ref{fig:Introduction1} compares the architecture of HQ-OV3D with prior works. The IMCV Proposal Generator employs Vision-Language Models (VLMs) to identify novel objects in 2D images, coupled with a dedicated cross-modality cross-validation pipeline that enforces geometric consistency between image and LiDAR modalities to enhance 3D pseudo-proposal quality. The ACA Denoiser leverages known-class annotations and incorporates a structure-aware super-category conditioning diffusion mechanism to refine candidate boxes for novel-class objects. This enables fine-grained optimization of proposals for unseen classes, resulting in high-quality open-vocabulary pseudo-labels.
%This framework effectively integrates the  structural and geometric priors extracted from base-class annotations, enabling more reliable pseudo-label generation for novel classes. It provides a solid data foundation for subsequent multi-modal modeling and cross-modal alignment.

%Our main contributions and innovations are summarized as follows. (1) We propose a \textit{Intra-Modality Cross-Validated Proposal Generator} module that addresses key limitations in existing OV-3D pseudo-label gerneration pipelines by integrating spatial localization and geometric reasoning under a dual-constraint strategy. A carefully designed cluster merging algorithm is employed to generate high-quality 3D proposals for novel classes. (2) We design a diffusion-based proposal refinement module, \textit{Annotated-Class Assisted Denoiser}, which incorporates structure-aware super-category conditioning. This module effectively refines novel-class proposals by leveraging only base-class supervision. (3) Experimental results show that training with pseudo-labels generated by HQ-OV3D achieves a 7.37\% mAP improvement on novel classes compared to the current state-of-the-art, demonstrating the superior quality of our pseudo-labels and the strong potential of HQ-OV3D as a robust foundation for open-vocabulary 3D detection.
Our main contributions and innovations are summarized as follows. (1) We propose a \textit{Intra-Modality Cross-Validated Proposal Generator}, a novel module designed to overcome the limitations of low-quality pseudo-labels in existing approaches through cross modality geometric constraint-based confidence modeling and adaptive cluster merging algorithm. (2) We design a diffusion-based proposal refinement module, \textit{Annotated-Class Assisted Denoiser}, effectively refines novel-class proposals by leveraging known-class annotations based on a diffusion framework. (3) Experimental results show that training 3D detector with pseudo-labels generated by HQ-OV3D achieves a \textbf{7.37\%} mAP improvement on novel classes compared to the current state-of-the-art, proving its dual efficacy as both a standalone open-vocabulary 3D detector and high-quality pseudo-label generator for existing open-vocabulary 3D detection pipeline.

\section{Related Work}
\subsection{3D Object Detection}

Conventional 3D object detection methods are mainly tailored for closed-set scenarios, relying on fully annotated data within fixed category definitions. LiDAR-based approaches are commonly categorized into point-based~\cite{yang2018ipod,yang2019std,shi2019pointrcnn,shi2020point,yang20203dssd,zhao2021point}, voxel-based~\cite{yan2018second,lang2019pointpillars,zhou2018voxelnet,yang2018pixor,deng2021voxel,yin2021center}, hybrid point-voxel~\cite{shi2020pv,shi2023pv,li2021lidar,liu2019point}, and range image-based methods~\cite{meyer2019lasernet,liang2020rangercnn,liang2021rangeioudet,fan2021rangedet}, each differing in spatial modeling, computational efficiency, and real-time performance. Recently, Transformer architectures have been widely adopted in 3D detection for their global modeling capabilities. Some methods embed Transformer~\cite{vaswani2017attention} modules into backbones or detection heads~\cite{sheng2021improving,mao2021voxel,pan20213d}, while others follow DETR-style designs with query-based modeling over multi-modal features~\cite{wang2022detr3d,li2024bevformer,bai2022transfusion,li2022unifying,chen2023futr3d}, achieving both strong performance and scalability.

Conventional closed-set 3D detection methods heavily rely on predefined categories and dense annotations, making them ill-suited for diverse real-world scenarios. They also struggle to recognize unseen categories during inference, exhibiting poor generalization.

\subsection{Open-Vocabulary Detection}

With the rise of large-scale VLMs, open-vocabulary object detection has become a prominent research direction. In the 2D domain, methods align region features with category embeddings by distilling semantic knowledge from CLIP~\cite{radford2021learning}. GLIP~\cite{li2022grounded} formulates detection as a phrase-matching task, while GroundingDINO~\cite{liu2024grounding} proposes a decoupled image-text fusion strategy. YOLO-World~\cite{cheng2024yolo} further integrates open-vocabulary modeling into the YOLO architecture for efficient detection.

Building upon the success of open-vocabulary paradigms in 2D vision, recent efforts have begun exploring their extension to 3D object detection. However, progress in OV-3D remains limited, particularly in LiDAR-based outdoor scenarios, due to the scarcity of large-scale point cloud–text pairs~\cite{luo2023kecor,luo2023exploring}. Early works such as OV-3Det~\cite{lu2023open} proposed lifting image-level labels generated by VLMs to 3D in indoor scenes. ImOV3D~\cite{yang2024imov3d} bypassed point clouds entirely via depth prediction, but its indoor-focused design limits applicability to outdoor settings. To improve robustness in outdoor environments, OpenSight~\cite{zhang2024opensight} incorporated spatiotemporal context and geometric priors, while CoDA~\cite{cao2023coda} introduced a semi-supervised framework with limited manual annotations. Find n’ Propagate~\cite{etchegaray2024find} explored dense proposal generation in frustum space, and OV-SCAN~\cite{chow2025ov} enhanced semantic alignment for better open-set recognition.

Despite these advances, most current methods rely heavily on heuristic projection or frustum-based search to generate 3D pseudo-labels for novel classes. These approaches often lack fine-grained localization and optimization mechanisms, leading to suboptimal spatial accuracy and limited quality of OV-3D pseudo-labels.

\section{Preliminaries}

In our setting of OV-3D, we define the base training dataset as $\mathcal{D}_B = \{\mathcal{L}, \mathcal{I}, \mathcal{B}_B\}$, where $\mathcal{L}$ denotes a set of raw point clouds, $\mathcal{I}$ represents corresponding multi-view images, and $\mathcal{B}_B$ is the set of 3D bounding boxes associated with base classes $\mathcal{C}^B$. The goal of OV-3D in our setting is to obtain a 3D detector $f(\cdot\,;\theta)$ trained with $\mathcal{D}_B$, capable of detecting objects from both base and novel class sets, i.e., $\mathcal{B}_B \cup \mathcal{B}_N$, despite novel classes $\mathcal{C}^N$ have no annotated samples and satisfying $\mathcal{C}^N \cap \mathcal{C}^B = \emptyset$.

\section{Method}

\begin{figure*}[t]
\centering
\includegraphics[width=0.85\textwidth]{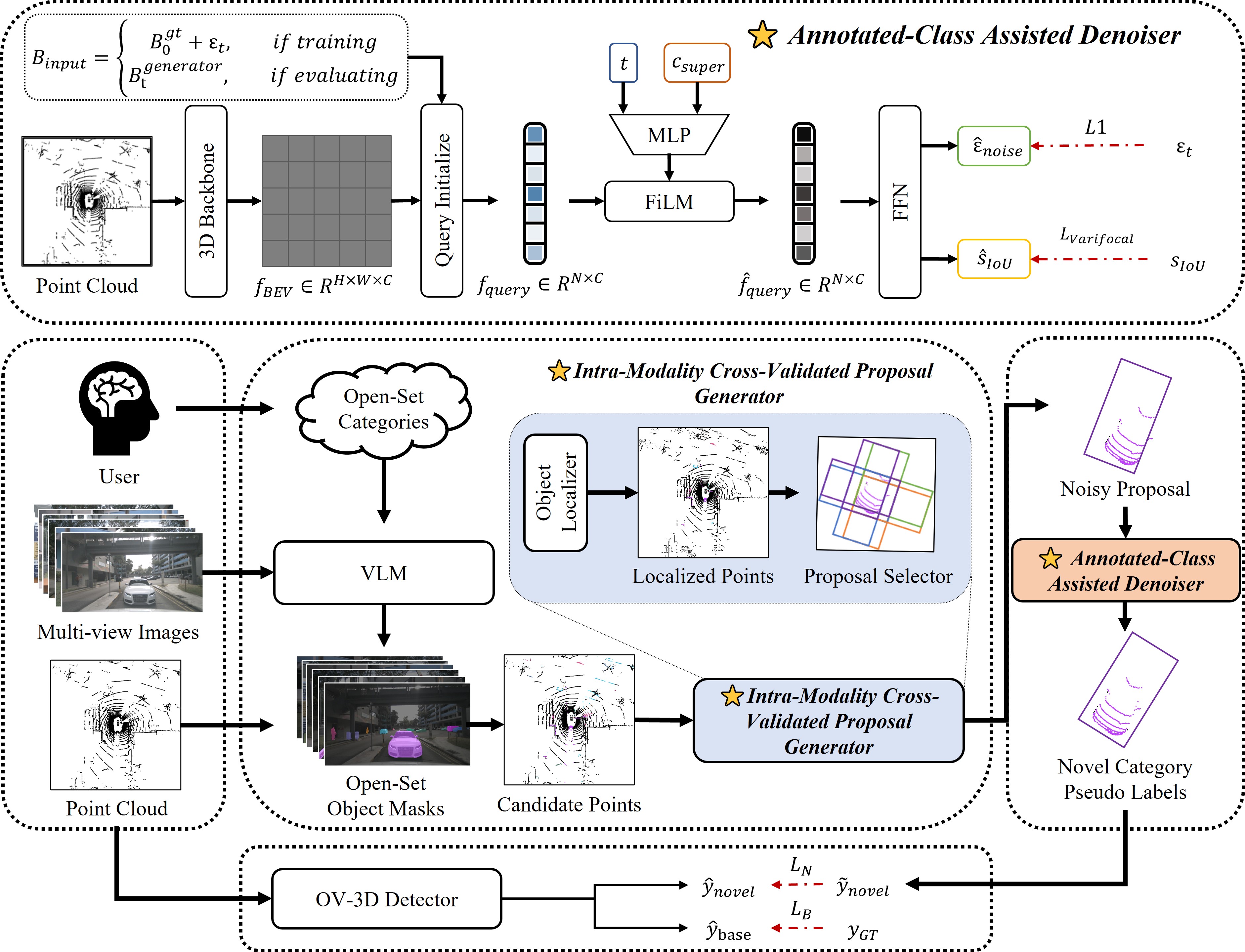}
\caption{Overview of the proposed \textbf{HQ-OV3D} framework. The pipeline consists of two main modules: the \textit{Intra-Modality Cross-Validated Proposal Generator}, which generates proposals for novel classes, and the \textit{Annotated-Class Assisted Denoiser}, which refines noisy proposals through a diffusion-based mechanism. Together, these modules enable high-quality pseudo-label generation for OV-3D.}
\label{fig:overview}
\end{figure*}

To overcome the inaccuracies in 3D proposals localization and geometric shape in existing heuristic-based projection and fitting pipeline, and fully leverage the available base-class GT annotations for enhancing the quality of novel-class proposals, we introduce a novel two-stage framework \textbf{HQ-OV3D} to improve box quality for open-vocabulary 3D object detection. Figure~\ref{fig:overview} illustrates the overall architecture of our framework.

In the first stage, we propose a \textit{Intra-Modality Cross-Validated Proposal Generator} (IMCV Proposal Generator) module that utilizes geometric consistency between image and LiDAR modalities to generate more complete, accurate 3D initial proposals for novel classes.
In the second stage, we propose a \textit{Annotated-Class Assisted Denoiser} (ACA Denoiser) module. Leveraging annotated base-class data and a denoising diffusion framework, this module refines the noisy initial proposals under zero-shot conditions, effectively transfers base classes box geometric info to novel classes, improves localization accuracy and geometric quality for novel classes detection.

\subsection{Intra-Modality Cross-Validated Proposal Generator}
Intra-Modality Cross-Validated Proposal Generator consists of three modules: Object Seeker, Object Localizer and Proposal Selector. 

\subsubsection{Object Seeker}

To discover novel-class objects from class set $\mathcal{C}^N$, Object Seeker module utilizes VLM to generate 2D detection boxes from $M$ camera views, resulting in a collection of detection boxes $\{B_{2D}^{(m,n)}\}_{m=1,n=1}^{M,N_m}$, where $B_{2D}^{(m,n)}$ denotes the $n$-th detected box from the $m$-th view. Each detected box is then fed into the SAM model~\cite{kirillov2023segment}to segment the foreground region within the box, producing a set of foreground masks $\{R_{\text{fg}}^{(m,n)}\}_{m=1,n=1}^{M,N_m}$.

%The masks are served as 2D semantic and spatial priors in the subsequent Object Localizer and Selector modules.

\subsubsection{Object Localizer}

\begin{algorithm}
\caption{Cluster Merging for Object Localizer}
\begin{algorithmic}[1]
\Require Cluster set $\mathcal{C} = \{c_1, c_2, \dots, c_N\}$, Geometry Confidence Score $S_{\text{geo}}$, threshold $\text{thresh}_{\text{dim}}$
\Ensure Final merged cluster $\hat{c}^{(m,n)}$, bounding box $\hat{b}^{(m,n)}$

\State Sort clusters $\mathcal{C}$ by $S_{\text{geo}}$ in descending order
\State Initialize merged cluster $c_{\text{merged}} \gets \mathcal{C}[1]$
\State Fit 3D box $b \gets \text{FitBox}(c_{\text{merged}})$
\State Compute initial $S_{\text{IoU}} \gets \text{ConsistencyScore}(b)$

\For{each cluster $c$ in $\mathcal{C}[2:]$}
    \State $c_{\text{temp}} \gets c_{\text{merged}} \cup c$
    \State Fit box $b_{\text{temp}} \gets \text{FitBox}(c_{\text{temp}})$
    \State Compute $S_{\text{IoU}}^{\text{temp}} \gets \text{ConsistencyScore}(b_{\text{temp}})$
    \If{$S_{\text{IoU}}^{\text{temp}} > S_{\text{IoU}}$ \textbf{and} $\text{BoxDimDiff}(b, b_{\text{temp}}) < \text{thresh}_{\text{dim}}$}
        \State $c_{\text{merged}} \gets c_{\text{temp}}$
        \State $b \gets b_{\text{temp}}$
        \State $S_{\text{IoU}} \gets S_{\text{IoU}}^{\text{temp}}$
    \EndIf
\EndFor

\State \Return $c_{\text{merged}}$ as $\hat{c}^{(m,n)}$, $b$ as $\hat{b}^{(m,n)}$
\end{algorithmic}
\end{algorithm}

%Existing heuristic-based  methods for generating proposals of novel classes typically rely on spatial constraints between image detection boxes and point cloud projections. These approaches extract object points by selecting point clouds that fall within the image detection box areas. However, such methods are limited by occlusion, sparse observations, and point cloud noise, which often lead to inaccurate proposal localization.

%To address this issue, we propose an Object Localizer module that enhances the integrity and accuracy of candidate point cloud through spatial confidence modeling and a greedy cluster merging algorithm.

The Object Localizer module identifies candidate 3D points using 2D foreground masks $\{R_{\text{fg}}^{(m,n)}\}_{m=1,n=1}^{M,N_m}$ as spatial constraints. To address the challenges of misaligned and missed points due to the depth ambiguous across 3D-to-2D, 2D masks boundaries inaccuracy and lidar sparsity, cross-modality cross-validated confidence modeling and greedy 3D cluster merging algorithm are dedicated designed.

Given the global point cloud $\mathcal{P}$ of the current frame and the set of foreground regions ${R}_{\text{fg}}$, we first project the point cloud into each camera view and extract the projected points within each foreground region:
\begin{equation}
\mathcal{P}_{\text{fg}}^{(m,n)} = \{p_i \in \mathcal{P} \mid \Pi_m(p_i) \in \mathcal{R}_{\text{fg}}^{(m,n)}\},
\end{equation}
where $\Pi_m(\cdot)$ denotes the projection function mapping 3D points to the image plane of the $m$-th camera.

To estimate the relevance of each point $p_i^{(m,n)} \in \mathcal{P}_{\text{fg}}^{(m,n)}$ to the object, we compute a \textit{Geometry Score} $s_{\text{geo}}^i$, which captures the normalized spatial deviation from the image center and point cloud range:
\begin{equation}
s_{\text{geo}}^i = \text{Normalized}\left(1 - \sqrt{\frac{(x_i')^2 + (y_i')^2 + (r_i')^2}{3}}\right),
\end{equation}
where $x_i'$ and $y_i'$ are the normalized offsets within the image box, and $r_i'$ is the normalized range value. Higher scores indicate that the point is closer to ego-vehicle center and detected object center in 2D image, indicating more likely to belong to the true object.

Next, we apply clustering to the candidate point clouds, resulting in a set of clusters:$\mathcal{C}^{(m,n)} = \{ \mathcal{C}_k^{(m,n)} \}_{k=1}^{K_{(m,n)}}$
We define a \textit{Geometry Confidence Score} for each cluster based on its aggregated point-level geometry scores:
\begin{equation}
S_{\text{geo}}^{(k,(m,n))} = \frac{\sum_{p_i \in \mathcal{C}_k^{(m,n)}} s_{\text{geo}}^i}{\sum_{p_j \in \mathcal{P}_{\text{fg}}^{(m,n)}} s_{\text{geo}}^j}.
\end{equation}

Each cluster $\mathcal{C}_k^{(m,n)}$ is then fitted with a minimal 3D bounding box $b_k^{(m,n)}$.
To assess its geometric consistency with the 2D detection, we compute an \textit{IoU Score} via the projected IoU between the 3D box and the image detection box:
\begin{equation}
S_{\text{IoU}}^{(k,(m,n))} = \text{IoU}(\text{Proj}_m(b_k^{(m,n)}), B_{2D}^{(m,n)}),
\end{equation}
where $\text{Proj}_m$ denotes the projection of the 3D box onto the image plane of camera $m$.

To further improve the completeness of the object region, we introduce a cluster merging algorithm (see Algorithm 1). Starting from the cluster with the highest $S_{\text{geo}}$, we iteratively merge other clusters in descending order of their confidence scores. After each merge, the 3D box is refitted, and the consistency score $S_{\text{IoU}}$ is updated. Merging is retained only if the IoU improves and the box dimensions remain within a predefined threshold $\text{thresh}_{\text{dim}}$. The final merged cluster $\hat{\mathcal{C}}^{(m,n)}$ and its 3D bounding box $\hat{b}^{(m,n)}$ serve as the output of the Object Localizer.

\subsubsection{Proposal Selector}
Although the Object Localizer provides high-confidence point clusters $\hat{\mathcal{C}}^{(m,n)}$, directly fitting box according to $\hat{\mathcal{C}}^{(m,n)}$ often yields suboptimal size and orientation estimates due to occlusion and sparsity of point clouds. Proposal Selector module is proposed to select the optimal object box from a set of high-coverage box candidates generated from $\hat{\mathcal{C}}^{(m,n)}$.

We first obtain the typical size prior $s_{\text{prior}}^c$ of the current object category in real-world scenarios using GPT-4. Guided by this prior, we then construct $N_b$ standard-size candidate boxes based on the initial 3D box $\hat{b}^{(m,n)}$, and adjust their centers to better align with the surrounding point cloud boundaries. For each size configuration, we apply $N_\theta$ discrete yaw angles to form the full candidate set:
\begin{equation}
\mathcal{B}_{\text{cand}} = \left\{ b^{(i,j)} = (c^i, s^i, \theta_j) \right\}_{i=1,j=1}^{N_b, N_\theta},
\end{equation}
where $c^i$ is the adjusted center, $s^i$ the candidate box size, and $\theta_j$ is the yaw angle.

To select the final object box $b_{\text{final}}$, we compute a total score $S_{\text{total}}^{(i,j)}$ for each candidate $b^{(i,j)}$ by combining the \textit{Point Cloud Coverage Score} and the \textit{IoU Score}:
\begin{equation}
S_{\text{pts}}^{(i,j)} = \frac{|\mathcal{P}_{\text{in}}|}{|\mathcal{P}_{\text{total}}|},
\end{equation}
\begin{equation}
S_{\text{total}}^{(i,j)} = \alpha_{\text{pts}} S_{\text{pts}}^{(i,j)} + \alpha_{\text{IoU}} S_{\text{IoU}}^{(i,j)}, \quad \alpha_{\text{pts}} + \alpha_{\text{IoU}} = 1,
\end{equation}
where $\mathcal{P}_{\text{in}}$ is the number of points inside the box, and $\mathcal{P}_{\text{total}}$ is the total number of points in the current cluster. The candidate with the highest total score is selected as the final object box $b_{\text{final}}$.

\begin{figure*}[t]
\centering
\includegraphics[width=0.85\textwidth]{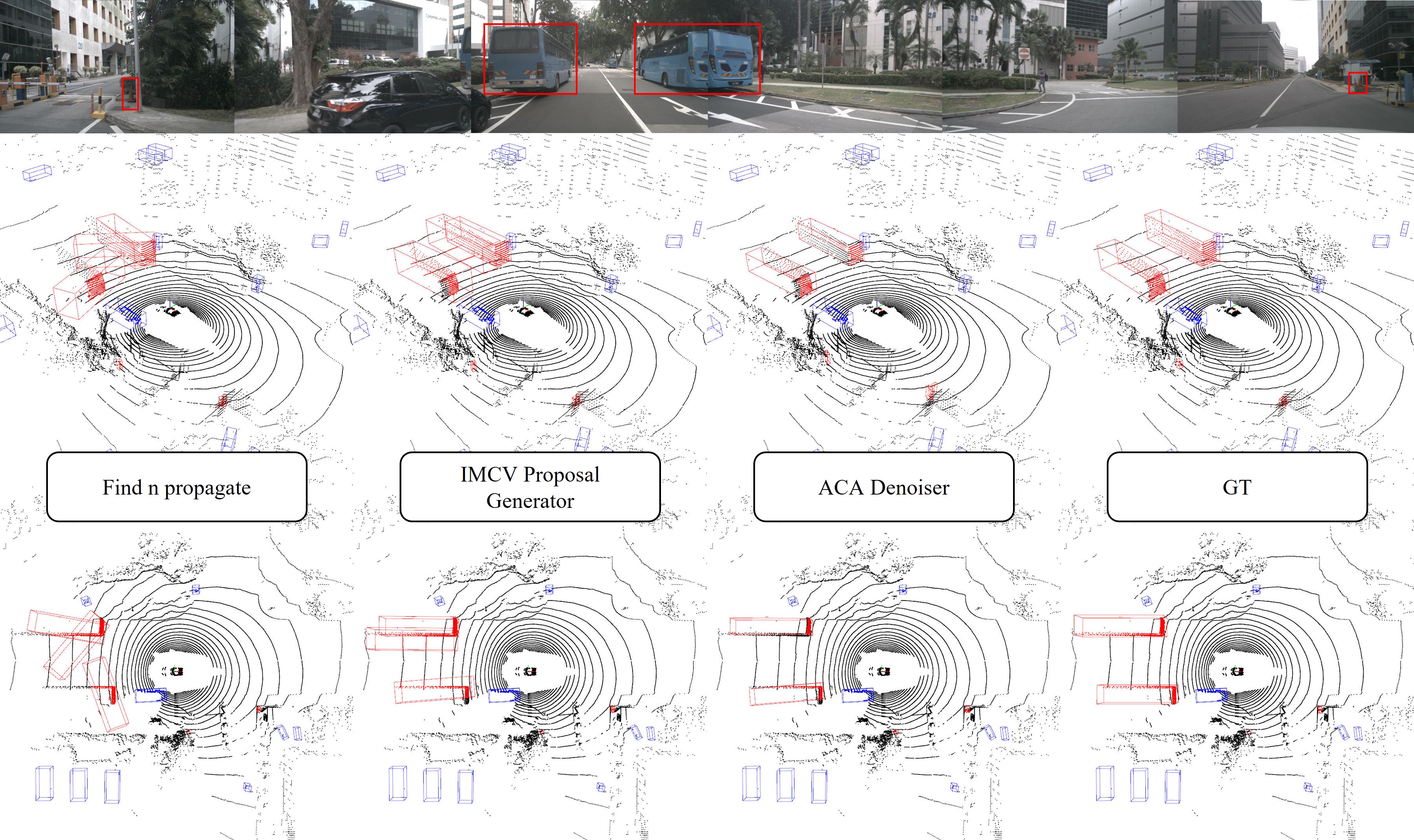}
\caption{This figure compares the novel-class pseudo-labels generated by our HQ-OV3D framework and Find n’ Propagate.  
Red boxes indicate objects from novel classes.}
\label{fig:visualize}
\end{figure*}

\subsection{Annotated-Class Assisted Denoiser}

%To further improve the quality and localization accuracy of initial proposals generated by the \textit{IMCV Proposal Generator}, we introduce the \textit{Annotated-Class Assisted Denoiser} (ACA Denoiser), a diffusion-based 3D box refinement module. 

While the IMCV Proposal Generator produces higher-quality initial proposals $b_{\text{final}}$, systematic deviations persist when compared to annotated boxes of known classes. We introduce the \textit{Annotated-Class Assisted Denoiser} (ACA Denoiser) to transfer known classes geometric prior to novel classes in a a diffusion-based 3D box refinement module.

\subsubsection{Diffusion-Based Refinement}

We model each proposal from IMCV Proposal Generator $b_{\text{final}}$ as a noisy observation $\hat{b}_t$ of a true object box $ b_0 $ perturbed by an unobservable systematic noise $\varepsilon_{\text{sys}}$:
\begin{equation}
\hat{b}_t = b_0 + \varepsilon_{\text{sys}}, \quad \varepsilon_{\text{sys}} \sim \mathcal{E}.
\end{equation}
This formulation reveals a key insight: proposal refinement can be viewed as an inverse denoising process. To this end, we adopt DDIM~\cite{song2020denoising} for iterative residual prediction, progressively refining proposals towards ground truth boxes. DDIM provides a deterministic sampling path, which maintains modeling flexibility while reducing iteration steps, yielding efficient and stable inference.

To ensure generalization to novel classes without supervision, we propose a \textit{Super Category Condition} strategy. Instead of using semantic labels, we guide the denoising process with structural similarity. Specifically, we map each category $c$ to a discrete super category index in a set $ \mathcal{C}_{\text{super}} $ based on category-wise geometric size priors $s_{\text{prior}}^c$, where each super category represents a group of object types with similar geometry. Each super category is embedded as a vector $e^{\text{super}}$ and used as a conditional input to modulate feature extraction and residual prediction.

At each denoising step $ t \in \{T, T-1, \ldots, 1\} $, the ACA Denoiser receives the current noisy box state $ x_t $ and predicts a residual:
\begin{equation}
\Delta b_t = (\Delta c_t, \Delta s_t, \Delta \theta_t),
\end{equation}
where $\Delta c_t \in \mathbb{R}^3$ is the center offset, $\Delta s_t \in \mathbb{R}^3$ the size residuals, and $\Delta \theta_t \in \mathbb{R}^2$ the yaw angle delta. The proposal state is updated as:
\begin{equation}
x_{t-1} = \alpha_t \cdot (x_t - \epsilon_\theta(x_t, t, c^{\text{super}})) + \beta_t \cdot z, \quad z \sim \mathcal{N}(0, I),
\end{equation}
where $\epsilon_\theta$ denotes the residual prediction network, $\alpha_t$ and $\beta_t$ are DDIM coefficients, and $c^{\text{super}}$ is the super category label.

At each step, the ACA Denoiser extracts the local BEV feature $ f_{\text{local}} \in \mathbb{R}^C $ at the proposal center $(x, y)$ and embeds the box parameters $b \in \mathbb{R}^7$ using a geometric encoder $\phi_{\text{box}}(\cdot)$. The final query representation is:
\begin{equation}
q = [f_{\text{local}} \,\|\, \phi_{\text{box}}(b)] \in \mathbb{R}^{2C},
\end{equation}
where $\|$ denotes concatenation. The query is fused with the global BEV feature map $F_{\text{global}} \in \mathbb{R}^{C \times H \times W}$ via cross-attention:
\begin{equation}
f_{\text{query}} = \text{CrossAttn}(q, F_{\text{global}}).
\end{equation}

We then concatenate the time-step embedding $e^t$ with the super category embedding $e^{\text{super}}$ and project them through an MLP to obtain a conditional modulation vector:
\begin{equation}
e^{\text{cond}} = \text{MLP}([e^t, e^{\text{super}}]),
\end{equation}
which is used to generate affine transformation parameters in a Feature-wise Linear Modulation (FiLM) manner:
\begin{equation}
\tilde{f} = \gamma \cdot f_{\text{query}} + \beta, \quad [\gamma, \beta] = \text{MLP}(e^{\text{cond}}).
\end{equation}
The modulated feature $\tilde{f}$ is finally passed to the prediction head to produce the residual $\Delta b_t$.

During training, we sample gaussian noise to perturb the ground truth box $ b_0^{\text{gt}}$ and generate a noisy input $ b_t^{\text{gt}}$. Although the actual error distribution may not strictly follow a gaussian distribution, gaussian noise provides a class-agnostic and general-purpose approximation, making it a reasonable surrogate.
The model is supervised to predict the residual using an L1 loss:
\begin{equation}
\mathcal{L}_{\text{res}} = \text{L1}(b_t^{\text{gt}} + \Delta b_t - b_0^{\text{gt}}).
\end{equation}

Through this structure-aware conditional denoising mechanism, ACA Denoiser learns a generalizable capability to recover clean box geometry from noisy proposals, without requiring any annotations for novel classes.

\subsubsection{IoU-Aware Confidence Score}

To improve the quality-awareness of the refined proposals $\hat{b}_0$, we introduce an IoU-Aware Confidence Score mechanism. A confidence branch is trained to predict the geometric quality of each proposal, supervised by its 3D IoU with the ground truth. We adopt Varifocal Loss~\cite{zhang2021varifocalnet}to encourage accurate quality estimation.

At inference, we combine the predicted geometric confidence with the VLM-provided category score to obtain a fused confidence score. This fusion balances geometric consistency and semantic relevance, enabling more accurate pseudo-label selection and ranking, especially for novel categories without ground-truth supervision.

\section{Experiments}
\begin{table}[t]
\centering
\scriptsize
\setlength{\tabcolsep}{2pt}
\renewcommand{\arraystretch}{1.1}
\begin{tabular}{@{}l|l|c c c c|c@{}}
\toprule
\textbf{Method}               & \textbf{VLM}   & \textbf{truck} & \textbf{bus} & \textbf{motor} & \textbf{cone} & \textbf{mAP} \\
\midrule
{Opensight~\cite{zhang2024opensight}} & GDINO  
    &  11.60 &  5.10 &  30.40 &  25.60 
    &  18.18 \\
{OV-SCAN~\cite{chow2025ov}} & GDINO  
    &  \textbf{30.30} &  22.00 &  39.80 &  39.60 
    &  32.93 \\
{Find n'~\cite{etchegaray2024find}} & GLIP  
    & 26.17 & 28.43 & 34.18 & 45.83 
    & 33.65 \\
\midrule

\multirow{2}{*}{\textbf{HQ-OV3D}}
    & GLIP  
    & 25.30 & 29.80 & 51.20 & 57.80 
    & \textbf{41.02} \\
    & GDINO 
    & 13.60 & \textbf{30.60} & \textbf{54.80} & \textbf{62.20} 
    & 40.30 \\
\bottomrule
\end{tabular}
\caption{Comparison of OV-3D methods on the nuScenes dataset. ``GDINO'' denote GroundingDINO.}
\label{tab:pseudo_label_training}
\end{table}

\begin{table*}[t]
\centering
\scriptsize
\setlength{\tabcolsep}{3pt}
\renewcommand{\arraystretch}{1.1}
\begin{tabular}{@{}l|l|cccccccccc|cc@{}}
\toprule
\multirow{2}{*}{\textbf{Method}}
  & \multirow{2}{*}{\textbf{VLM}}
  & \multicolumn{10}{c|}{\textbf{Novel}}
  & \multicolumn{2}{c}{\textbf{Overall}} \\
& 
  & \textbf{car} & \textbf{const} & \textbf{trai} & \textbf{barr} & \textbf{bic} 
  & \textbf{ped} & \textbf{truck} & \textbf{bus} & \textbf{motor} & \textbf{cone}
  & \textbf{mAP} & \textbf{NDS} \\
\midrule

\multirow{2}{*}{OpenSight~\cite{zhang2024opensight}}
    & OWL   
    & 14.20 & 1.80 & 0.00 & 0.10 &  5.20 
    & 19.40 &  2.50 &  4.00 &  5.50 &  1.10 
    &  5.40 & 12.40 \\

    & Detic 
    & 15.10 & 2.10 & 0.00 & 0.10 &  6.20 
    & 21.10 &  2.90 &  4.20 &  6.10 &  0.80 
    &  5.80 & 12.70 \\

\midrule
\multirow{3}{*}{Find\,n'~\cite{etchegaray2024find}}
    & GLIP
    & 19.20 & 3.70 & 0.10 & 4.10 & 33.60 
    & 18.70 &  7.90 &  9.20 & 32.10 & 20.50 
    & 14.89 & 18.52 \\
    
    & GDINO 
    & 18.90 & 0.00 & 0.00 & 0.30 & \textbf{35.80} 
    & 39.90 &  2.50 &  9.10 & \textbf{33.90} & 45.70 
    & 18.61 & 18.86 \\

    & GT2D  
    & 16.09 & 3.26 & 0.37 & 34.28 & 38.22 
    & 39.03 &  7.41 &  5.54 & 30.33 & 56.45 
    & 23.10 & 22.83 \\

\midrule
\multirow{3}{*}{\textbf{HQ-OV3D}}
    & GLIP  
    & \textbf{21.00} & \textbf{5.50} & \textbf{0.30} & \textbf{5.40} 
    & 31.70 & 19.90 & \textbf{10.80} & 12.10 
    & 31.30 & 21.70 & 15.96 & 21.37 \\
    
    & GDINO 
    & 19.90 & 0.00 & 0.00 &  0.50 & 35.50
    & \textbf{42.10} &  3.60 & \textbf{13.20} & 33.80 & \textbf{47.30} 
    & \textbf{19.60} & \textbf{21.68} \\

    & GT2D  
    & 28.10 & 11.50 & 3.20 & 56.20 & 50.50 
    & 61.60 & 18.20 & 16.50 & 43.90 & 65.40 
    & 35.50 & 31.81 \\
\bottomrule
\end{tabular}
\caption{Comparison of OV-3D methods on the nuScenes dataset with 10 novel classes.}
\label{tab:generator}
\end{table*}
\subsection{Experimental Setup}
\paragraph{Dataset.}
We evaluate our HQ-OV3D approach on the nuScenes dataset~\cite{caesar2020nuscenes}, which includes 1,000 driving sequences with approximately 1.4 million annotated 3D boxes. The training, validation, and test splits contain 700, 150, and 150 sequences, respectively. Each scene includes synchronized 1600$\times$900 RGB images from six cameras covering a full 360-degree view.

\paragraph{Open-vocabulary Settings.}
We designate six categories as Base classes: \textit{car, construction vehicle, trailer, barrier, bicycle, pedestrian}, and treat the remaining four categories as Novel classes. This class split follows the approach in~\cite{yin2021center}, where categories are grouped based on geometric similarity to ensure reasonable task difficulty and shared structural characteristics between base and novel classes. 

\paragraph{Evaluation Metrics.}
We adopt the official nuScenes evaluation metrics, including mean Average Precision (mAP) and nuScenes Detection Score (NDS). 

\paragraph{Training Setup.}
The \textit{ACA Denoiser} module is trained for 20 epochs on two NVIDIA A100 GPUs with a batch size of 4, using the AdamW optimizer with an initial learning rate of 0.0001 and a weight decay of 0.01. To incorporate structural priors, all object categories are grouped into five super categories based on geometric size and structural similarity: super category 0 contains: \textit{car}, \textit{truck}, \textit{construction vehicle}; super category 1 contains: \textit{bus}, \textit{trailer}; super category 2 contains: \textit{barrier}; super category 3 contains: \textit{motorcycle}, \textit{bicycle}; super category 4 contains: \textit{pedestrian}, \textit{traffic cone}.

\subsection{Main Results}

To systematically evaluate the quality of pseudo-labels generated by our method for novel classes, we directly use them as supervision signals to train a Transfusion~\cite{bai2022transfusion} detector on the nuScenes dataset, and compare our approach with mainstream OV-3D methods. Our training does not involve any modality alignment, focusing solely on assessing the effectiveness of the pseudo-labels themselves. As shown in Table~\ref{tab:pseudo_label_training}, our method outperforms existing approaches in terms of mAP of novel classes, verifying the higher quality and stronger supervision capability of the pseudo-labels produced by our framework. In particular, our method surpasses the current state-of-the-art method Find n’ Propagate~\cite{etchegaray2024find} approach by a significant margin of 7.37\%  in mAP, demonstrating the superiority of our framework.

To evaluate the performance of our \textit{IMCV Proposal Generator}, we conduct experiments on the nuScenes dataset by treating all ten categories as novel classes, without any supervision. As shown in Table~\ref{tab:generator}, our method outperforms OpenSight~\cite{zhang2024opensight} and Find n’ Propagate~\cite{etchegaray2024find} across various metrics, demonstrating its ability to generate higher-quality proposals.

\subsection{Ablation Study}
\begin{table}[t]
\centering
\scriptsize
\setlength{\tabcolsep}{3pt}
\renewcommand{\arraystretch}{1.1}
\begin{tabular}{@{}l|l|c c c c|c@{}}
\toprule
\textbf{Method}               & \textbf{VLM}   & \textbf{truck} & \textbf{bus} & \textbf{motor} & \textbf{cone} & \textbf{mAP} \\
\midrule
Find\,n'~\cite{etchegaray2024find}                      & GLIP     &  7.90 &  9.20 & 32.10 & 20.50 & 17.43 \\
\textbf{HQ-OV3D*}                     & GLIP     & 10.80 & 12.10 & 31.30 & 21.70 & 18.98 \\
\textbf{HQ-OV3D}            & GLIP     & \textbf{14.30} & \textbf{18.30} & \textbf{37.30} & \textbf{22.10} & \textbf{23.00} \\
\midrule
Find\,n'~\cite{etchegaray2024find}                      & GDINO    &  2.50 &  9.10 & 33.90 & 45.70 & 22.80 \\
\textbf{HQ-OV3D*}                     & GDINO    &  3.60 & 13.20 & 33.80 & 47.30 & 24.48 \\
\textbf{HQ-OV3D}            & GDINO    &  \textbf{4.90} & \textbf{19.80} & \textbf{37.90} & \textbf{48.50} & \textbf{27.78} \\
\midrule
Find\,n'~\cite{etchegaray2024find}                      & GT2D     &  7.40 &  5.50 & 30.30 & 56.40 & 24.90 \\
\textbf{HQ-OV3D*}                     & GT2D     & 18.20 & 16.50 & 43.90 & 65.40 & 36.00 \\
\textbf{HQ-OV3D}            & GT2D     & \textbf{27.80} & \textbf{26.40} & \textbf{51.20} & \textbf{66.80} & \textbf{43.05} \\
\bottomrule
\end{tabular}
\caption{Ablation study of \textit{ACA Denoiser}, * indicates that the ACA Denoiser is not used to refine pseudo-labels.}
\label{tab:ablation_Denoiser}
\end{table}

\paragraph{ACA Denoiser}
To verify the effectiveness of the proposed ACA Denoiser module in proposal refinement, we conduct comparative experiments on the nuScenes dataset. Table~\ref{tab:ablation_Denoiser} presents the evaluation results of pseudo-label quality under different settings. The results show that incorporating the ACA Denoiser effectively improves the accuracy of pseudo-labels for novel categories, achieving up to a 4.02\% increase in mAP.  These results confirm that our proposed module effectively mitigates the systematic bias introduced by heuristic-based pesudo-label generation, further enhancing the precision and usability of the resulting pseudo-labels.

\paragraph{Diffusion-based Refinement}
To specifically evaluate the effectiveness of the diffusion architecture in our ACA Denoiser module, we design two representative baselines for comparison. All models use GroundingDINO to generate 2D detections.

\textit{Heuristic-based  Temporal Refinement.} This method enhances recall via object tracking and interpolation across frames, but its effectiveness heavily depends on the quality of initial detections. Inaccurate early predictions disrupt association across frames, leading to unreliable interpolation. As shown in Table~\ref{tab:ablation_Diffusion_model}, unstable pseudo-labels may even lead to performance degradation after temporal refinement.

\textit{Transformer-based Refinement.} This model is also trained with base-class annotations as supervision. During training, Gaussian noise is added to the GT boxes, and the model learns to predict the residual to recover the original boxes. Unlike the diffusion architecture, it directly predicts the residual after extracting fused features via cross attention. As shown in Table~\ref{tab:ablation_Diffusion_model}, the Transformer-based model struggles to generalize its refinement ability to unseen classes compared to the diffusion model, indicating the diffusion framework offers better generalization and robustness when refining pseudo-labels for novel classes.

\begin{table}[t]
\centering
\scriptsize
\setlength{\tabcolsep}{3pt}
\renewcommand{\arraystretch}{1.1}
\begin{tabular}{@{}l|c c c c|c@{}}
\toprule
\textbf{Method}               & \textbf{truck} & \textbf{bus} & \textbf{motor} & \textbf{cone} & \textbf{mAP} \\
\midrule
HQ-OV3D (IMCV Proposal Generator)                  & 3.60 & 13.20 & 33.80 & 47.30 & 24.48 \\
Heuristic-based  Temporal Refinement               & 4.60 & 12.40 & 30.50 & 21.70 & 23.68 \\
Transformer-based Refinement      & 4.70 & 14.80 & 35.00 & 44.10 & 24.65 \\
\textbf{HQ-OV3D*}       & \textbf{5.40} & 18.40 & 36.70 & 46.30 & 26.70 \\
\textbf{HQ-OV3D}       & 4.90 & \textbf{19.80} & \textbf{37.90} & \textbf{48.50} & \textbf{27.78} \\
\bottomrule
\end{tabular}
\caption{Ablation study of \textit{Diffusion-based Refinement} and \textit{Super Category Condition}, * indicates that the Super Category Condition is not used.}
\label{tab:ablation_Diffusion_model}
\end{table}

\paragraph{Super Category Condition}
To validate the effectiveness of Super Category Condition in enhancing the model's generalization ability, we compare the ACA Denoiser model using only timestep modulation with the version that incorporates both super category and timestep modulation. As shown in Table~\ref{tab:ablation_Diffusion_model}, the addition of super category conditioning shows a clear potential to improve the model’s generalization to novel classes.

\section{Conclusion and Discussion}

In this work, we propose \textbf{HQ-OV3D}, a high-quality open-vocabulary 3D object detection framework designed to generate and refine accurate pseudo-labels for novel categories. The framework consists of two key components: an \textit{Intra-Modality Cross-Validated} (IMCV) Proposal Generator, which exploits cross-modality geometric consistency to produce high-quality initial 3D proposals, and an \textit{Annotated-Class Assisted} (ACA) Denoiser, which progressively refines proposals through a DDIM-based denoising process by leveraging geometric priors from annotated categories. This design effectively mitigates projection IoU underestimation caused by partial visibility. Experimental results show that, under a fully zero-shot setting, HQ-OV3D achieves a \textbf{7.37\%} mAP improvement on novel classes compared to state-of-art methods, demonstrating the superior quality and reliability of the generated pseudo-labels. 

Notably, replacing VLM-based 2D detections with ground-truth boxes yields up to an \textbf{80\%} performance improvement, clearly indicating that in the future work, stronger 2D VLM-based detection model can further significantly enhance our HQ-OV3D performance. 

\section{Acknowledgments}
This work was supported by the National Key R\&D Program of China, Project “Development of Large Model Technology and Scenario Library Construction for Autonomous Driving Data Closed-Loop” (Grant No. 2024YFB2505501).This work was also supported by the Guangxi Major Science and Technology Project “Research and Industrialization of Key Technologies for High-Intelligence and Cost-Effective Urban Leading Driving” (Grant No. AA24206054).

% The preparation of the \LaTeX{} and Bib\TeX{} files that implement these instructions was supported by Schlumberger Palo Alto Research, AT\&T Bell Laboratories, Morgan Kaufmann Publishers, The Live Oak Press, LLC, and AAAI Press. Bibliography style changes were added by Sunil Issar. \verb+\+pubnote was added by J. Scott Penberthy. George Ferguson added support for printing the AAAI copyright slug. Additional changes to aaai2026.sty and aaai2026.bst have been made by Francisco Cruz and Marc Pujol-Gonzalez.

% \bigskip
% \noindent Thank you for reading these instructions carefully. We look forward to receiving your electronic files!

\bibliography{aaai2026}
\newpage
\appendix
\section{Appendix}
\subsection{Ablation Study of IoU-Aware Confidence Score}
From Table~\ref{tab:confidence_weight_ablation}, we observe that incorporating the IoU-Aware Confidence Score leads to a moderate improvement in mAP, with the best performance achieved at a weight of 0.6. This demonstrates that, even when the model is trained only on base categories, the IoU Score can effectively assess the refinement quality of proposals for novel categories. However, assigning excessive weight to the IoU Score degrades performance, as it captures geometric quality but lacks semantic confidence, which is provided by the VLM. A balanced combination of the two is therefore essential.

\begin{figure}[htbp]
\centering
\includegraphics[width=0.45\textwidth]{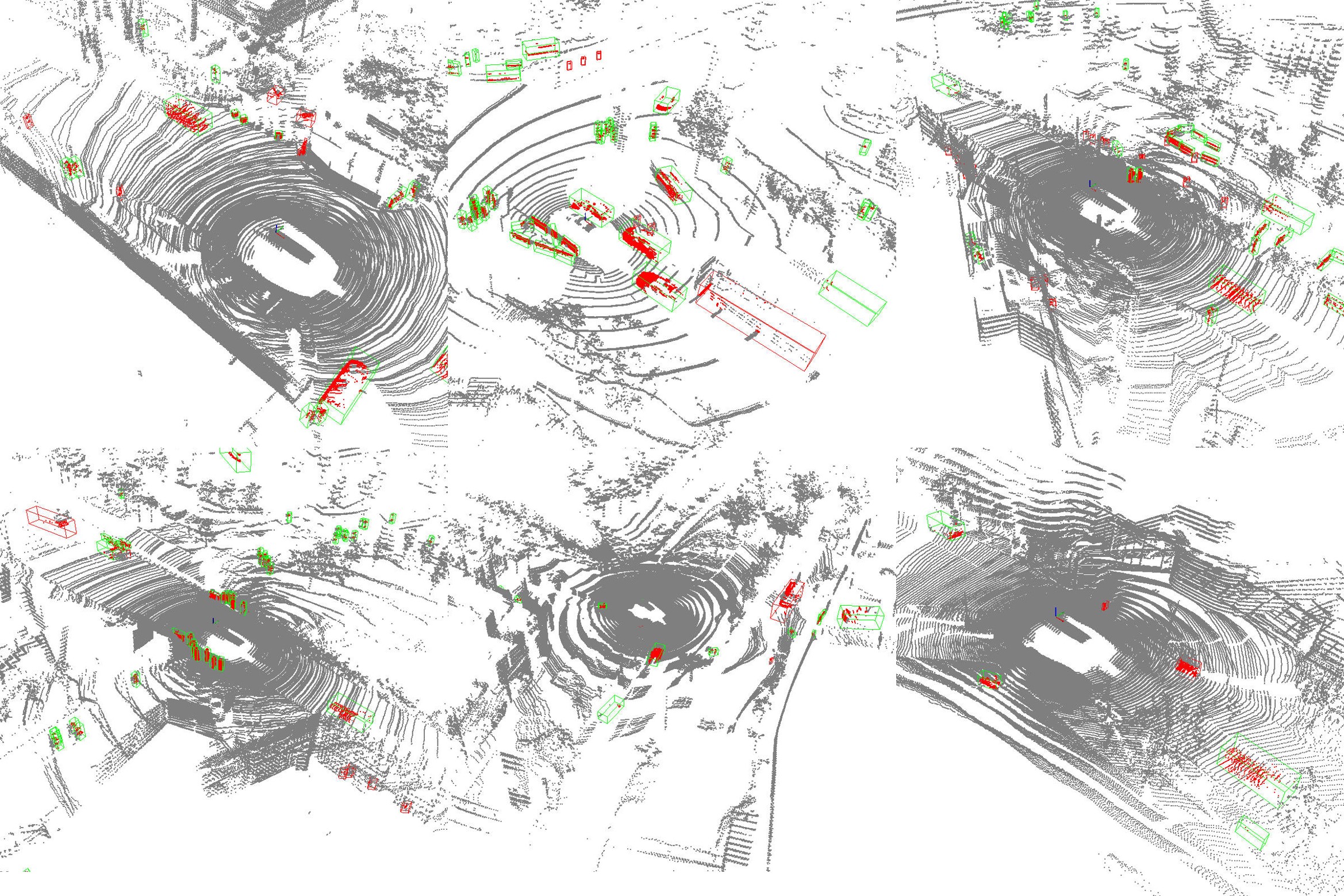}
\caption{Detection results of the downstream detector trained with pseudo labels generated by HQ-OV3D. Green boxes denote objects of base classes, while red boxes indicate novel classes objects.}
\label{fig:visualize_supp}
\end{figure}

\begin{table}[htbp]
\centering
\scriptsize
\setlength{\tabcolsep}{4pt}
\renewcommand{\arraystretch}{1.1}
\begin{tabular}{@{}c c c c@{}}
\toprule
\textbf{IoU Confidence Weight} & \textbf{VLM Confidence Weight} & \textbf{mAP} \\
\midrule
0.0 & 1.0 & 26.88 \\
0.2 & 0.8 & 27.08 \\
0.4 & 0.6 & 27.38 \\
\textbf{0.6} & \textbf{0.4} & \textbf{27.78} \\
0.8 & 0.2 & 27.55 \\
1.0 & 0.0 & 21.60 \\
\bottomrule
\end{tabular}
\caption{Effect of confidence weight balancing between IoU score and VLM score on pseudo-label quality. The VLM used is GroundingDINO~\cite{liu2024grounding}. Results evaluated on the nuScenes dataset.}
\label{tab:confidence_weight_ablation}
\end{table}

% Example: Insert a figure
% Uncomment and modify the following lines to add your own figures:
% \begin{figure}[h]
% \centering
% \includegraphics[width=0.9\columnwidth]{your-figure-name}
% \caption{Your figure caption here.}
% \label{fig:supp1}
% \end{figure}
\subsection{Implementation Details}

\subsubsection{Experimental Setup}
The Transfusion detector~\cite{bai2022transfusion} is trained on two NVIDIA A100 GPUs with a batch size of 4 for 20 epochs. The optimizer settings are kept consistent with those of the Annotated-Class Assisted (ACA) Denoiser, using AdamW with a learning rate of 0.0001 and weight decay of 0.01.

\subsubsection{IMCV Proposal Generator}
In the \textit{Proposal Selector} component of our method, we incorporate class-specific dimension priors to guide proposal refinement. The dimension priors for each object category, obtained from GPT-4, are summarized in Table~\ref{tab:dimension_priors}. These priors are also used in the \textit{Object Localizer} module, where we define a threshold dimension parameter, $\text{thresh}_{\text{dim}}$, for each class. Specifically, the$\text{thresh}_{\text{dim}}^{\text{c}}$of each class is set to:
${
 \text{thresh}_{\text{dim}}^{\text{c}} = 1.2 \times \text{dim}_{\text{prior}}^{\text{c}}.
}$
This threshold is used to associate candidate points within a plausible spatial extent of the object.

As part of the pre-clustering process in the \textit{Object Localizer}, we adopt DBSCAN to group spatially adjacent candidate points. Specifically, we set the DBSCAN parameters to $\text{eps} = 0.50$ and $\text{min}_\text{samples} = 1$. This setting ensures that only very close points are clustered together, while isolated candidates are still retained as individual clusters. Such configuration is intentionally designed to avoid overly aggressive merging, thereby preserving localization fidelity and ensuring that potentially valid proposals are not prematurely filtered out.

\begin{table}[h]
\centering
\begin{tabular}{@{}l|ccc@{}}
\toprule
\textbf{Category} & \textbf{Length} & \textbf{Width} & \textbf{Height} \\
\midrule
Car & 4.63 & 1.97 & 1.74 \\
Truck & 6.93 & 2.51 & 2.84 \\
Construction Vehicle & 6.37 & 2.85 & 3.19 \\
Bus & 10.50 & 2.94 & 3.47 \\
Trailer & 12.29 & 2.90 & 3.87 \\
Barrier & 0.50 & 2.53 & 0.98 \\
Motorcycle & 2.11 & 0.77 & 1.47 \\
Bicycle & 1.70 & 0.60 & 1.28 \\
Pedestrian & 0.73 & 0.67 & 1.77 \\
Traffic Cone & 0.41 & 0.41 & 1.07 \\
\bottomrule
\end{tabular}
\caption{Dimension priors for each object category obtained from GPT-4.}
\label{tab:dimension_priors}
\end{table}

\subsection{Visualization}
Figure~\ref{fig:visualize_supp} illustrates the detection performance of a downstream detector trained jointly with pseudo labels for novel categories generated by HQ-OV3D and ground-truth annotations for base categories. The pseudo-labeled data enables the detector to generalize beyond the base categories and identify novel objects to a certain extent.

% ----------- Supplementary Content Ends Here -----------

% References and End of Paper
% These lines must be placed at the end of your paper

\end{document}